%% file: acl_latex.tex
\pdfoutput=1

\documentclass[11pt]{article}

\usepackage[preprint]{acl}

\usepackage{times}
\usepackage{latexsym}
\usepackage{longtable}
\usepackage[T1]{fontenc}

\usepackage[utf8]{inputenc}
\usepackage{amsmath}
\usepackage{algorithm}
\usepackage{algorithmic}
\usepackage{microtype}
\usepackage{booktabs} 
\usepackage{inconsolata}
\usepackage{enumitem}
\usepackage{graphicx}
\usepackage{array}

\title{Debt Collection Negotiations with Large Language Models: An Evaluation System and Optimizing Decision Making with Multi-Agent}

\author{
Xiaofeng Wang\textsuperscript{1,2}, Zhixin Zhang\textsuperscript{2}\footnotemark[1], Jinguang Zheng\textsuperscript{2}, Yiming Ai\textsuperscript{1,2}, Rui Wang\textsuperscript{1}\thanks{Corresponding Authors} \\
\textsuperscript{1}Shanghai Jiao Tong University \\
\textsuperscript{2}Ant Group \\
\{banyedy, wangrui12\}@sjtu.edu.cn \\
\{zhangzhixin.zzx, zhengjinguang.zhen\}@antgroup.com \\
}

\begin{document}
\maketitle
\begin{abstract}
Debt collection negotiations (DCN) are vital for managing non-performing loans (NPLs) and reducing creditor losses. Traditional methods are labor-intensive, while large language models (LLMs) offer promising automation potential. However, prior systems lacked dynamic negotiation and real-time decision-making capabilities. This paper explores LLMs in automating DCN and proposes a novel evaluation framework with 13 metrics across 4 aspects. Our experiments reveal that LLMs tend to over-concede compared to human negotiators. To address this, we propose the \textbf{M}ulti-\textbf{A}gent \textbf{De}bt \textbf{N}egotiation \textbf{(MADeN)} framework, incorporating planning and judging modules to improve decision rationality. We also apply post-training techniques, including DPO with rejection sampling, to optimize performance. Our studies provide valuable insights for practitioners and researchers seeking to enhance efficiency and outcomes in this domain.
\end{abstract}

\input{latex/sections/1-introduction}
\input{latex/sections/2-data}

\input{latex/sections/3-benchmark}

\input{latex/sections/4-Results}
\input{latex/sections/5-methods}

\input{latex/sections/7-conclusion}

\section*{Limitations}
We study the performance and improvement methods of large models in debt collection negotiations. To simplify the research process and capture key negotiation points, we reduce the debtor’s financial information to variables like assets, average income, and average expenses. However, real-world financial situations are more complex, involving factors like cash flow issues and income fluctuations during repayment. Future work should involve more detailed simulations of debtor information and comparisons with manually simulated debtors. Additionally, due to time constraints, our creditor Multi-agent framework is relatively simple. In practical applications, stricter classification processes in \textit{planning} and more standardized methods in \textit{judging} are needed. We aim to integrate existing decision models to further optimize decision-making in the dialogue.
\section*{Ethical Considerations} \label{EC}
Our study does not disclose any real client information. The acquisition of the source data was subject to strict approval by a major internet financial institution, and the process was continuously supervised by relevant personnel. All debt-related data are processed and replaced with synthetic values, and names are substituted with the pseudonym ``Zhang San''. For the debt reasons extracted from collection dialogues, we strictly anonymize any sensitive details and provide generalized summaries, ensuring that no specific information is involved. Each final data entry underwent rigorous manual verification. Additionally, the methodology proposed in this paper is exploratory and based on simulation for research purposes. Prior to its application in real-world debt collection involving actual individuals, it will undergo more rigorous validation and approval processes.

We conducted annotation tasks in two areas: scoring for the Dialogue Soundness metric (Section~\ref{sec:eval}) and comparison with the human baseline as a debt collector (Section~\ref{sec:res}). Five graduate students with engineering backgrounds and two professionals with financial industry experience participated (They are all from China, as our study focuses on the Chinese language). All annotators involved in our study have signed a disclaimer acknowledging the terms and conditions associated with their participation. They were recruited through campus forums and the internal annotation program of the company. The annotation tasks did not involve any sensitive information and posed no risk. Compensation was provided according to the time spent on each task.
\section*{Acknowledgments}
This work was supported by Ant Group.

\bibliography{latex/custom}

\appendix
\input{latex/sections/6-related-work}

\input{latex/sections/Appendix-data-task}
\input{latex/sections/Appendix-metrics}
\input{latex/sections/Appendix-try}
\end{document}

%% file: latex/sections/1-introduction.tex
\section{Introduction}


Finance, as a negotiation-intensive field, involves the distribution and exchange of financial interests, requiring a higher level of understanding of information and rational decision-making \citep{Chan2006TheGN,Thompson1997TheMA}. Due to various personal financial issues, a large volume of non-performing loans (NPLs) arises each year across banks and financial companies, with debtors often being unable to repay their debts after prolonged overdue periods \citep{Ozili2019NonPerformingLA}. Negotiation and mediation are necessary to resolve their credit issues and minimize the losses for financial institutions (creditors) \citep{Firanda2021DebtCO}. Traditionally, the debt collection process has been labor-intensive, and data shows that in China, 3,800 financial institutions rely on outsourced specialized collection agencies to help recover non-performing assets \citep{Tang2018DebtCO}.

Previous automated debt collection dialogue models \citep{Floatbot2023GenerativeAI,Yahiya2024AutomatedDR} were primarily based on fixed-format notifications, where the models lacked communication and negotiation capabilities. Additionally, automated decision models \citep{Sancarlos2023TowardsAD,Jankowski2024DebtCM} related to changes in repayment strategies could not be directly integrated into the dialogue and were unable to update decisions in real time based on the debtor’s information provided during the conversation. A pressing need exists for novel approaches to automate \textbf{debt collection negotiations (DCN)}.

The rapid development of large language models (LLMs) \citep{Vicuna, LLaMA} and agent-based interactions \citep{luo2025llmpoweredmultiagentautomatedcrypto,chai2025flexquantelasticquantizationframework} built upon them has made it possible. Through emerging functions such as planning \citep{huang2024understanding}, reasoning \cite{aksitov2023rest}, and reflection \citep{renze2024self}, these models are now able to assist humans in completing more complex tasks. In this paper, we aim to explore the potential of using LLMs to support AI agents in performing this unexplored task. And firstly, it is crucial to develop a method to evaluate the performance in conducting DCN.


To develop a benchmark, the primary challenge lies in constructing a suitable dataset. In Section~\ref{sec:data}, to ensure both privacy and data validity, we utilized CTGAN \citep{ctgan} to generate synthetic data based on debt records from a leading financial technology company \footnote{The synthetic data generated in this work is publicly available.}. 
We supplement the debtor’s personal financial data through extraction and construction. Finally, we constructed a dataset containing 975 debt records. Based on this information, we provide a complete definition for DCN and the LLM-based negotiation process in Section~\ref{sec:task}.

\begin{figure*}[htbp]
\vspace{-0.1in}
  \centering
  \includegraphics[width=1\textwidth]{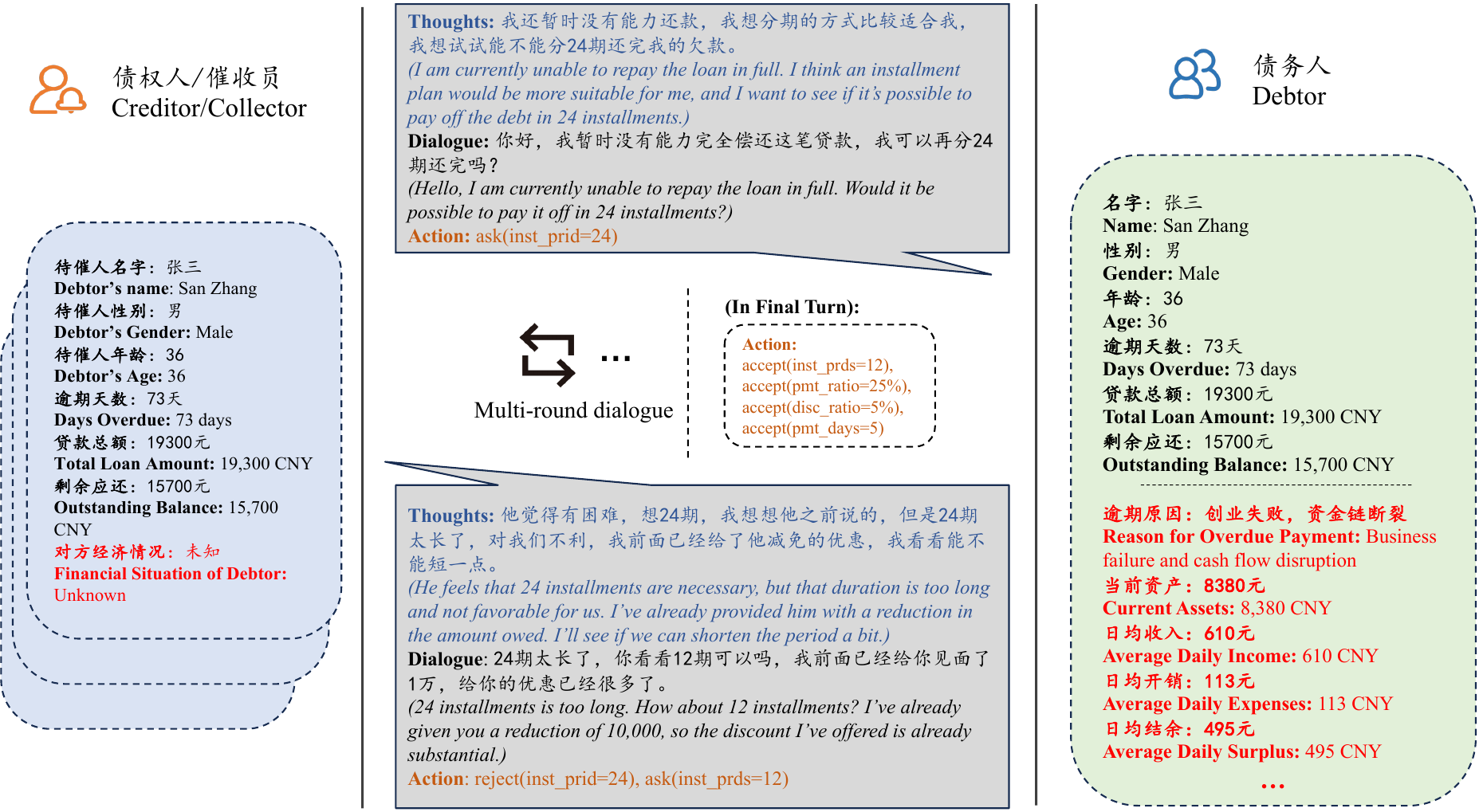}  
  \vspace{-0.15in}
  \caption{
An Example of a Debt Collection Negotiation (DCN). On the left and right sides are the information cards representing the data controlled by the debtor and the creditor, respectively. The black text represents the \textbf{basic debt information}, while the red text represents the \textbf{debtor’s personal financial information}. In the center, we demonstrate the use of LLM-based agents to simulate the dialogue. Each time, both the debtor and the creditor output a set of (Thoughts, Dialogue, Action). \textcolor[RGB]{184,96,41}{\textbf{Thoughts}} refers to their internal thought process, visible only to themselves; \textbf{Dialogue} represents the conversation in natural language; and \textcolor[RGB]{56,84,146}{\textbf{Action}} refers to the specific activities represented in a formal language within the dialogue. Each negotiation consists of multiple rounds of such interactions, ultimately leading to the negotiation outcome. The English text was automatically translated using Google Translate.}
\vspace{-0.1in}
\label{img:pipeline}
\end{figure*}

To comprehensively evaluate DCN, in Section~\ref{sec:eval} we proposed a holistic assessment framework encompassing \textbf{10} specific metrics in 4 aspects and \textbf{3} comprehensive index to thoroughly evaluate the negotiation process and its outcomes. From the perspective of the negotiation process, we evaluate the completeness and soundness of the dialogues. Regarding negotiation outcomes, we evaluate two key aspects: for creditors, we focus on debt recovery rate and collection efficiency, while for debtors, we assess financial health by predicting future asset changes based on negotiation results and individual financial data. The indices are introduced to integrate the opposing relationship between creditor’s interests and debtor's financial health.

In Section~\ref{sec:res}, we tested the performance of LLMs and found that they are unable to make appropriate decisions based on the debtor’s financial condition and are more likely to make unsuitable \textit{concessions} than human beings. This may result from the models’ excessive focus on harmony and agreement, leading debt collectors to overlook the rationality of decisions. To address this, inspired by the work of MetaGPT~\citep{Hong2023MetaGPTMP}, we designed an LLM-based \textbf{M}ulti-\textbf{A}gent \textbf{De}bt \textbf{N}egotiation \textbf{(MADeN)} framework for DCN in Section~\ref{sec:framework}. In this framework, we enhanced the basic \textbf{Communicating} agent with two additional modules: \textbf{(1) Planning}, where the LLM agent designs a rough decision framework and outlines the potential outcomes based on the debtor’s initial reasons and demands; \textbf{(2) Judging}, which evaluates the rationality of each action and provides optimization suggestions. Our method improves the comprehensive collection index by nearly \textbf{10\%}.
 
In addition, we attempted to use the DPO post-training method include \citep{rafailov2024directpreferenceoptimizationlanguage} with reject sampling \citep{liu2024statisticalrejectionsamplingimproves} to align the debt collector’s focus on recovery rate and efficiency in Section~\ref{sec:post}. On the Qwen2.5-7B \citep{qwen2.5} model, we observed improvements across various metrics.

Our contributions are summarized as follows:

\begin{itemize}[leftmargin=15px]
    
    \item We proposed a synthetic debt dataset and a comprehensive framework for evaluating LLM performance in debt collection negotiations (DCN), using 13 metrics to assess both the negotiation process and outcomes, enabling the testing and evaluation of different models.
    \item Our testing of mainstream LLMs on this task revealed that the models tend to make decisions with more unreasonable concessions compared to humans.
    \item We developed an multi-agent framework for DCN, incorporating two key modules to improve negotiation outcomes. Additionally, we explored post-training the model through rejection sampling on multi-agent data, which also enhanced the model’s performances.
    
\end{itemize}


%% file: latex/sections/2-data.tex
\section{Data Collection}\label{sec:data}

Our data is primarily divided into two parts, as shown in Figure~\ref{img:pipeline}. The basic debt information is known to both the debtor and the creditor, while the debtor’s personal financial data is not accessible to the creditor. We now explain how each of these two data components was collected.

\subsection{Basic Debt Information}

The basic debt data primarily consists of personal information and debt-related information. We sampled from \textit{real debt data} provided by the financial company mentioned in introduction. To ensure privacy compliance, we used \textbf{CTGAN} \citep{ctgan} to generate synthetic data~\footnote{Please refer to our Ethical Considerations.}. We categorized the data by gender, overdue days and loan amount. Then we sampled it to match the distribution patterns of the original data. 

\subsection{Debtor’s personal financial data}

Debtor’s personal financial data collection involved two main components: \textbf{textual reasons for overdue} and \textbf{numerical financial information}. The reasons for overdue payments were extracted from real dialogue data and assigned to different categories based on their real \textit{distribution}. For numerical financial data, we simplified complex personal data into components such as total assets, average daily income, expenses, and surplus. Since this data is typically unavailable, we used a linear model with Gaussian noise, based on historical data correlations, to estimate these values. 

Finally, we collected \textbf{975} debt records, with \textbf{390} records placed in the test set and the remaining \textbf{585} records in the training set (The subsequent evaluations are conducted on the test set). Details of the debtor category distribution can be found in Appendix~\ref{Distribution}.

\section{Task Formulation} \label{sec:task}

\begin{algorithm}[!htb]
    \caption{\label{alg:1}Debt Collection Negotiation Process}
    \label{alg:2}
    \begin{algorithmic}
        \STATE \textbf{Initialize:} Action Set $S_A$, Basic Debt Information $I_b$, Personal Financial Information $I_p$, Agent \text{Creditor}, Agent \text{Debtor}, Maximum Turns $t_m$, Negotiation Dimensions Set $S_R$, Negotiation Result Dictionary $D$
        
        \STATE {$\text{Creditor} \gets \text{Creditor}(I_b, S_A)$}
        \STATE {$\text{Debtor} \gets \text{Debtor}(I_b, I_p, S_A)$}
        \STATE {$t \gets 0$}
        \STATE {$D \gets \{\}$} 
        
        \FOR{$t < t_m$}
            \STATE $ A_c, \text{Dialogue}_c \gets \text{Creditor.generate}$
            
            \STATE $\text{Debtor} \gets \text{Debtor}(A_c, \text{Dialogue}_c)$
            \STATE $A_d, \text{Dialogue}_d \gets \text{Debtor.generate}$
            \IF{$A_d == \texttt{accept}$} 
                \STATE $D[A_d.key] \gets A_d.value $ 

            \ENDIF
            \IF {$D$ covers $S_R$}
                \RETURN $D$
            \ENDIF   
            \STATE $\text{Creditor} \gets \text{Creditor}(A_d, \text{Dialogue}_d)$
            \STATE {$t \gets t + 1$}
        \ENDFOR
        
        \RETURN None
    \end{algorithmic}
\end{algorithm}

\subsection{Definition and Objectives}\label{obj}

\begin{table*}[ht]
\centering

\caption{\label{dimdes}Debt Collection Negotiation Dimensions}
\vspace{-0.1in}
    \setlength{\tabcolsep}{3.5mm}{
    \resizebox{\textwidth}{!}{%
        \begin{tabular}{lll}
        \toprule
        \textbf{Dimension} & \textbf{Range} & \textbf{Description} \\
        \midrule
        Discount Ratio  & 0 - 30\% & The portion of debt waived by the creditor to ease repayment. \\
        Immediate Payment Ratio & 5\% - 50\% & The portion of debt that must be repaid immediately, typically at least 5\%. \\
        Immediate Payment Time & 1 - 14 (days) & A grace period of up to 14 days for the debtor to make the immediate repayment. \\
        Installment Periods  & 3 - 24 (months) & The duration for repaying the remaining debt in installments.\\
        \bottomrule
        \end{tabular}
        \vspace{-0.1in}
    }}
    \vspace{-0.1in}
\end{table*}

Debt collection negotiations (DCN) refers to negotiations initiated by creditors to recover outstanding debts and restore the debtor’s credit, due to the debtor’s inability to repay on time because of personal financial issues. The measures for negotiating the resolution of non-performing loans generally include deferral, debt forgiveness, collateralization, conversion, and installment payments \citep{DFRatings2019,Lankao2023}. Among these, deferral, debt forgiveness, and installment payments are the most commonly used. We have distilled them into four dimensions: Discount Ratio, Immediate Payment Ratio, Immediate Payment Time and Installment Periods\footnote{Refer to \url{https://www.boc.cn/bcservice/bc3/bc31/201203/t20120331_1767028.html} for the calculation of installment interest.}. Table~\ref{dimdes} presents the range of values and a brief description of each dimension, and the detailed explanations are provided in Appendix~\ref{sec:dim}. Through negotiations on these four aspects, the goal of both parties is to reach a \textit{mutually acceptable outcome} that allows the debtor to resolve their outstanding debt in a manageable way.


\subsection{Future Economic Predictions for Debtors}

\begin{figure*}[htbp]
  \centering
  \includegraphics[width=\textwidth]{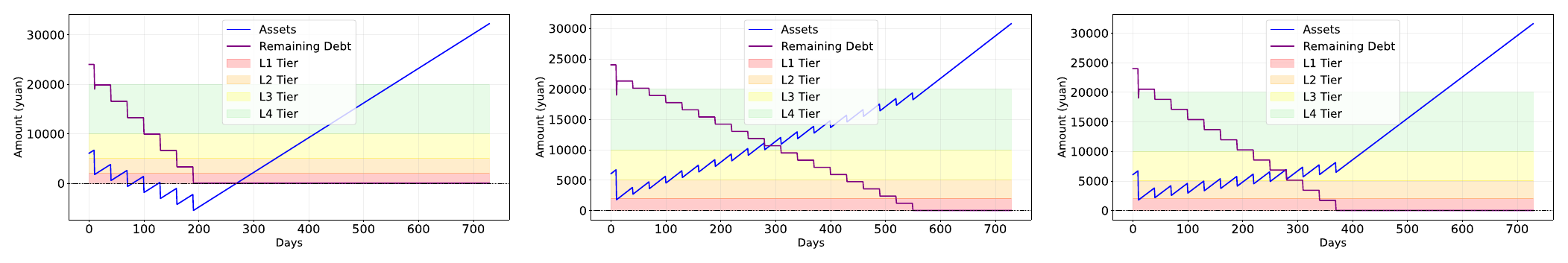}  
  \vspace{-0.3in}
  \caption{\label{image:pred}
The future trajectories of the debtor’s remaining assets and outstanding debt under three installment plans (6, 12, and 18 months from left to right) are shown, with all other variables held constant. The 6-month plan causes the debtor’s assets to fall \textbf{below zero}, making repayment impossible. In contrast, the 12-month and 18-month plans maintain a healthy asset level, though the 18-month plan significantly \textbf{reduces recovery efficiency}. The 12-month plan is the most balanced solution. Different background colors represent five difficulty tiers, with Tier 1 being the most challenging. The specific ranges and descriptions of the tiers are provided in Appendix~\ref{app:diff_cat}.}
\vspace{-0.1in}
\end{figure*}

After obtaining the negotiation results and integrating them with the debtor’s current financial model, we can project changes in their assets and remaining debt over the next \textit{two years}. Figure~\ref{image:pred} shows one debtor’s economic trajectory under three installment scenarios. In one scenario, the debtor’s assets fall into negative values, indicating a failed negotiation. In another, a too lenient installment plan reduces recovery efficiency. These scenarios provide a basis for evaluating negotiation outcomes, which will be discussed in Section~\ref{sec:eval}.

\subsection{Negotiation Process}

As shown is Figure~\ref{img:pipeline}, our negotiation process is a variant of the bargaining process designed by \citet{xia2024measuringbargainingabilitiesllms}. To formally articulate the negotiation between agents, we define the relevant concepts and variables in Table  ~\ref{tab:debt_variable_app}. A brief pseudo code of the process is Algorithm~\ref{alg:1}.

In the action set, \textit{“ask”}, \textit{“reject”} and \textit{“accept”} represent three different operations for each negotiation dimension. After several rounds of negotiation and discussion, the debtor and the collector can be considered to have reached an agreement when consensus \textit{(“accept”)} is achieved on all 4 negotiation objectives.

%% file: latex/sections/3-benchmark.tex
\begin{figure}[htbp]
\vspace{-0.1in}
  \centering
  \includegraphics[width=0.48\textwidth]{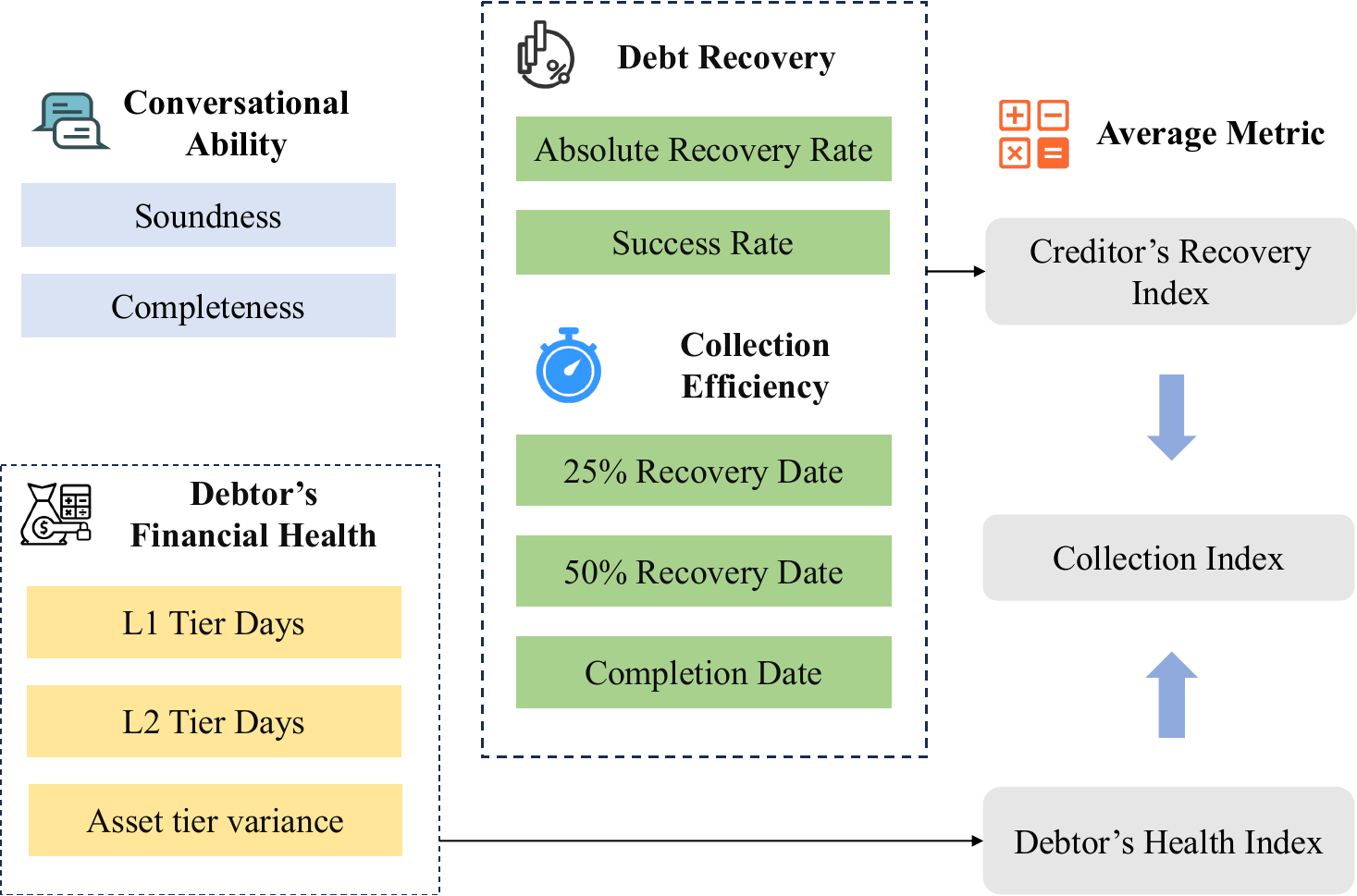}  
  \vspace{-0.1in}
  \caption{\label{image:eval}
 Evaluation system of DCN.}
\vspace{-0.1in}
\label{img:metric}
\end{figure}

\section{Evaluation System}\label{sec:eval}

Different from traditional negotiation evaluations, we argue that the DCN task requires a more comprehensive assessment framework. As illustrated in Figure ~\ref{image:eval}, we developed a evaluation system based on four aspects and extended several average metrics for a comprehensive assessment. 


\subsection{Segmented Evaluation Metrics}

In this section, we provide a general overview of the \textbf{10 metrics} across the four segmented aspects. Detailed descriptions, the evaluation process, and calculation formulas are further discussed in Appendix~\ref{app:metric}.

\textbf{Conversational Ability (§\ref{app:me_conv}).} Conversational ability is crucial in negotiation processes for effective communication and mutual understanding. We evaluate it using two metrics: \textit{(i) Dialogue Soundness}\textbf{ (DS)} is assessed on a five-point scale, measuring the fluency, naturalness and coherency of responses; \textit{(ii) Dialogue Completeness}\textbf{ (DC)} is an automated metric that evaluates whether four objectives are all addressed during the dialogue.

\textbf{Debt Recovery (§\ref{app:me_rec}).} In debt collection, the primary goal is to recover as much debt as possible. We evaluate this using two key metrics: \textit{(i) Success Recovery Rate} \textbf{(SR)} measures the proportion of samples where repayment can be successfully completed, based on the debtor’s future ability to meet repayment goals. \textit{(ii) Recovery Rate}\textbf{ (RR)} reflects the portion of the debt that has been successfully recovered by the creditor, calculated as the average recovery ratio across all test samples.

\textbf{Collection Efficiency (§\ref{app:me_col}).} Collection efficiency refers to how quickly a debtor can repay their debt. We monitor the timing of repayments using three key metrics: \textit{(i) 25\% Recovery Date}\textbf{ (QRD)} is the estimated date when the debtor has completed 25\% of the debt repayment, with earlier dates indicating quicker repayment; (\textit{ii) 50\% Recovery Date} \textbf{(HRD)} marks the completion of 50\% of the repayment, offering insight into the debtor’s ongoing repayment ability. \textit{(iii) The Completion Date} \textbf{(CD)} is the date when the debtor has fully repaid their debt, with a shorter completion date indicating a faster recovery process.

\textbf{Debtor’s Financial Health (§\ref{app:me_hea}).} The debtor’s financial health plays a critical role in successful debt recovery. It affects both the debtor’s ability to repay and the speed at which repayment occurs. We assess financial health using three metrics: \textit{(i) L1 Tier Days} \textbf{(L1D)} tracks the number of days the debtor remains in the most difficult financial tier (L1), with longer durations indicating higher risk of default; \textit{(ii) L2 Tier Days} \textbf{(L2D)} similarly tracks the days in the second most difficult financial tier (L2), which still reflects financial strain; \textit{(iii) Asset Tier Variance} \textbf{(ATV)} captures the variance in the debtor’s asset tier over a year, providing insight into the stability of their financial condition. 

\subsection{Comprehensive Indices}

We find that the indicators for recovery and efficiency are often \textit{\textbf{inversely related}} to the debtor’s financial condition in debt collection. To balance these conflicting objectives, we introduce three average metrics (detailed description and calculation process can be found in Appendic~\ref{app:me_ave}):

\textbf{(i) Creditor’s Recovery Index (CRI):} CRI is the \textit{weighted average} of five indicators from Debt Recovery and Collection Efficiency. It reflects an evaluation of the overall collection process by debt collectors, disregarding debtor-related factors. A higher value is more favorable to the creditor.

\textbf{(ii) Debtor’s Health Index (DHI):} DHI is the \textit{weighted average} of three indicators from Debtor’s Financial Health. It assesses the financial well-being of the debtor throughout the repayment process, with a higher value indicating a greater probability of the debtor adhering to the repayment plan.

\textbf{(iii) Comprehensive Collection Index (CCI):} CCI is the \textit{harmonic mean} of CRI and DHI. It provides a comprehensive evaluation of the negotiation outcome, where a higher value signifies the maximization of debt recovery and efficiency while ensuring the debtor’s financial health.

%% file: latex/sections/4-Results.tex
\section{Experiments and Results} \label{sec:res}

\begin{table*}[ht]
\vspace{-0.1in}
    \centering
    \caption{\label{img:mainresult}The performances of some models as debt collectors (~\textsuperscript{*} denotes the second-best performance).}
    \vspace{-0.1in}
    \setlength{\tabcolsep}{3.5mm}{
    \resizebox{\textwidth}{!}{%
    \begin{tabular}{l|cc|cc|ccc|ccc|cccc}
        \toprule
         & \multicolumn{2}{c}{\textbf{Conversation}} & \multicolumn{2}{c}{\textbf{Debt Recovery}} & \multicolumn{3}{c}{\textbf{Collection Efficiency}} & \multicolumn{3}{c}{\textbf{Debtor’s Financial Health}} & \multicolumn{3}{c}{\textbf{Average Metrics}}\\
        \cmidrule(lr){2-3} \cmidrule(lr){4-5} \cmidrule(lr){6-8} \cmidrule(lr){9-11} \cmidrule(lr){12-14} 
        \textbf{Model} & \textbf{DC}$ \uparrow $ & \textbf{DS}$ \uparrow $ & \textbf{SR}$ \uparrow $ & \textbf{RR(\%)}$ \uparrow $ & \textbf{QRD}$ \downarrow $ & \textbf{HRD}$ \downarrow $ & \textbf{CD}$ \downarrow $ & \textbf{L1D}$ \downarrow $ & \textbf{L2D}$ \downarrow $& \textbf{ATV}$ \downarrow $ & \textbf{CRI}$ \uparrow $ & \textbf{DHI}$ \uparrow $ & \textbf{CCI}$ \uparrow $\\
        \midrule
        Qwen-2.5-7B  & 0.94 & 4.57 & 0.98 & 87.15 & 46.04 & 214.04 & 436.84 & \textbf{2.82} & 79.46 & \textbf{0.84\textsuperscript{*}} & 0.732 & \textbf{0.793} & 0.743\\
        Qwen-2.5-14B & 0.94 & 4.60 & 0.96 & 89.62 & 28.60 & 154.60 & 358.80 & 6.30 & 79.82 & 0.88 & 0.793 & 0.613 & 0.749\\
        Qwen-2.5-72B & 0.96 & 4.75 & 0.98 & 88.50 & 36.98 & 185.18 & 404.98 & 3.76 & \textbf{78.84\textsuperscript{*}} & \textbf{0.83} & 0.764 & \textbf{0.767\textsuperscript{*}} & 0.764\\
        LLaMa-3-8B  & 0.91 & 3.64 & \textbf{1.00} & 89.01 & 51.36 & 184.02 & 399.02 & 3.38 & 88.02 & 0.87 & 0.756 & 0.713 & 0.747\\
        LLaMa-3-70B & 0.87 & 3.94 & 0.98 & 92.24 & 36.72 & 157.32 & 371.12 & 4.50 & 79.56 & 0.87 & 0.792 & 0.695 & 0.771\\
        GPT-4o & \textbf{1.00} & 4.65 & \textbf{1.00} & 95.76 & 27.00 & \textbf{128.40\textsuperscript{*}} & \textbf{297.20\textsuperscript{*}} & 6.18 & 85.18 & 0.90 & 0.844 & 0.580 & 0.774\\
        GPT-4o-mini & 0.99 & 4.61 & 0.96 & \textbf{96.32\textsuperscript{*}} & 31.60 & 131.20 & 312.00 & 6.30 & 84.08 & 0.89 & 0.836 & 0.589 & 0.771\\
        o1-mini & \textbf{1.00} & 4.68 & 0.94 & 94.61 & 29.52 & 140.52 & 352.52 & 5.58 & 83.80 & 0.89 & 0.807 & 0.619 & 0.760\\
        Doubao-pro & 0.98 & \textbf{4.91\textsuperscript{*}} & 0.96 & 93.11 & \textbf{21.22\textsuperscript{*}} & 143.02 & 365.22 & 5.98 & 83.68 & 0.89 & 0.814 & 0.603 & 0.760\\
        Claude-3.5 & \textbf{1.00} & 4.59 & 0.98 & 93.30 & 34.92 & 140.52 & 312.32 & \textbf{3.32\textsuperscript{*}} & 87.30 & 0.89 & 0.816 & 0.698 & \textbf{0.789\textsuperscript{*}}\\
        MiniMax & \textbf{1.00} & 4.75 & 0.96 & 92.77 & 38.66 & 167.66 & 401.26 & 7.12 & \textbf{76.44} & 0.88 & 0.776 & 0.591 & 0.730\\
        SenseChat & \textbf{1.00} & 4.70 & 0.98 & 89.28 & 34.56 & 155.76 & 354.56 & 5.24 & 81.14 & 0.87 & 0.791 & 0.661 & 0.761\\
        Deepseek-V3 & \textbf{1.00} & 4.85 & 0.99 & 91.65 & 28.40 & 141.20 & 313.60 & 5.42 & 83.82 & 0.89 & \textbf{0.818\textsuperscript{*}} & 0.625 & 0.771\\
        Deepseek-R1 & 0.98 & 4.81 & 0.98 & 93.10 & 37.72 & 146.32 & 348.12 & 5.68 & 83.94 & 0.88 & 0.802 & 0.624 & 0.759\\
        \midrule
        Human & \textbf{1.00} & \textbf{4.93} & \textbf{1.00} & \textbf{98.50} & \textbf{16.73} & \textbf{119.49} & \textbf{260.90} & 3.81 & 78.49 & 0.86 & \textbf{0.870} & 0.736 & \textbf{0.840}\\
        \bottomrule
    \end{tabular}%
 }}
 \label{tab:mainresults}
     \vspace{-10pt}
\end{table*}

In this section, we report the implementation details and the benchmark performances of several well-known LLMs in the DCN task on our dataset. 

\subsection{Implementation Details}


For \textit{open-source models}, such as Qwen series, we use their respective \textbf{chat} versions. For \textit{api-based models}, we aim to select the latest and most advanced versions available, the \textit{inference models} such as o1-mini \citep{openai2025o1} and DeepSeek-R1 \citep{deepseekai2025deepseekr1incentivizingreasoningcapability} are also included. The list of LLMs used is provided in Appendix~\ref{app:models}. The human baseline was derived from the average results of benchmark tasks completed by two finance professionals with relevant backgrounds.

Since our task focuses on Chinese, we chose one of the best open-source models currently available in the Chinese language domain: Qwen2.5-70B model to represent the \textbf{debtor}, while using different models for the \textbf{creditor} in order to compare their performance. The results of different models as the debtor are also presented in Appendix~\ref{sec:model_deb}. 

For both sides, we employed the Chain of Thought (CoT) approach \citep{wei2023chainofthoughtpromptingelicitsreasoning}, providing the model with instructions for the DCN task and a specified format for dialogue generation, which consisted of \textit{“Thought”}, \textit{“Dialogue”} and \textit{“Action” }in each interaction. The prompts are detailed in Appendix~\ref{app:prompts}.

\subsection{Benchmark Results}


The comparison of performance across different models is clearly illustrated in Table~\ref{img:mainresult}.

\textbf{LLMs perform well in terms of basic interaction format and overall dialogue capabilities.} From the perspectives of dialogue completeness (DC) and soundness (DS), we find that the models effectively cover all negotiation objectives. The dialogue content is generally reasonable, aligns with the set objectives, and shows little difference from the human baseline. Specifically, the Chinese-based model outperforms the English-based model in terms of dialogue soundness for our task.

\textbf{However, from the perspective of the negotiation outcomes, the performance of the LLMs was subpar and did not align well with requirements.} Observing the Comprehensive Collection Index (CCI), we found that the model’s overall evaluation result deviates from human-level performance by more than 0.05. This discrepancy might stem from the fact that the negotiation outcomes are numerical, making it challenging to align numerical-related requirements through prompt-based methods. 

\textbf{Most models tend to offer more generous concessions to debtors, both in repayment ratios and deadlines.} These concessions are crucial because they directly affect the financial company’s asset losses, a point emphasized in the prompt. However, as shown in the table, all models except for the GPT series have repayment ratios below 95\%, meaning they did not fully follow the prompt’s guidelines. In addition, the large models show lower collection efficiency compared to human benchmarks. For example, the time taken to recover 25\% of the debt is 2-3 times longer than the human baseline. This suggests the models give debtors more time to repay, rather than encouraging earlier repayment. Some models, like GPT-4o, come close to human-level efficiency, but this is at the cost of worsening the debtor’s financial situation. The average minimum repayment days for these models are twice as long as the human level, showing that they \textit{struggle to adapt} to the debtor’s real circumstances. This could be due to the models \textit{misjudging the debtor’s financial situation or choosing easier solutions to reach an agreement}.

\textbf{The collection results achieved by the model do not hold the debtor’s financial health, despite providing considerable room in terms of recovery and efficiency. }We found that, with the exception of the Qwen-2.5 model, the Debt Health Index (DHI) for all other models was below the human-level threshold. Considering the concessions offered to the debtor during the collection process, these results suggest that the model did not provide \textit{targeted debt resolution solutions} during the negotiation process.

\textbf{Non-inference models may be more suitable for this task compared to inference models.} By comparing the performance of the inference models o1-mini and Deepseek-R1 with their non-inference counterparts, gpt-4o-mini and Deepseek-V3, we observed a notable decline in the performance of the inference models across multiple metrics, particularly in collection efficiency.

%% file: latex/sections/5-methods.tex
\begin{figure*}[t]
\vspace{-0.35in}
  \centering
  \includegraphics[width=0.87\textwidth]{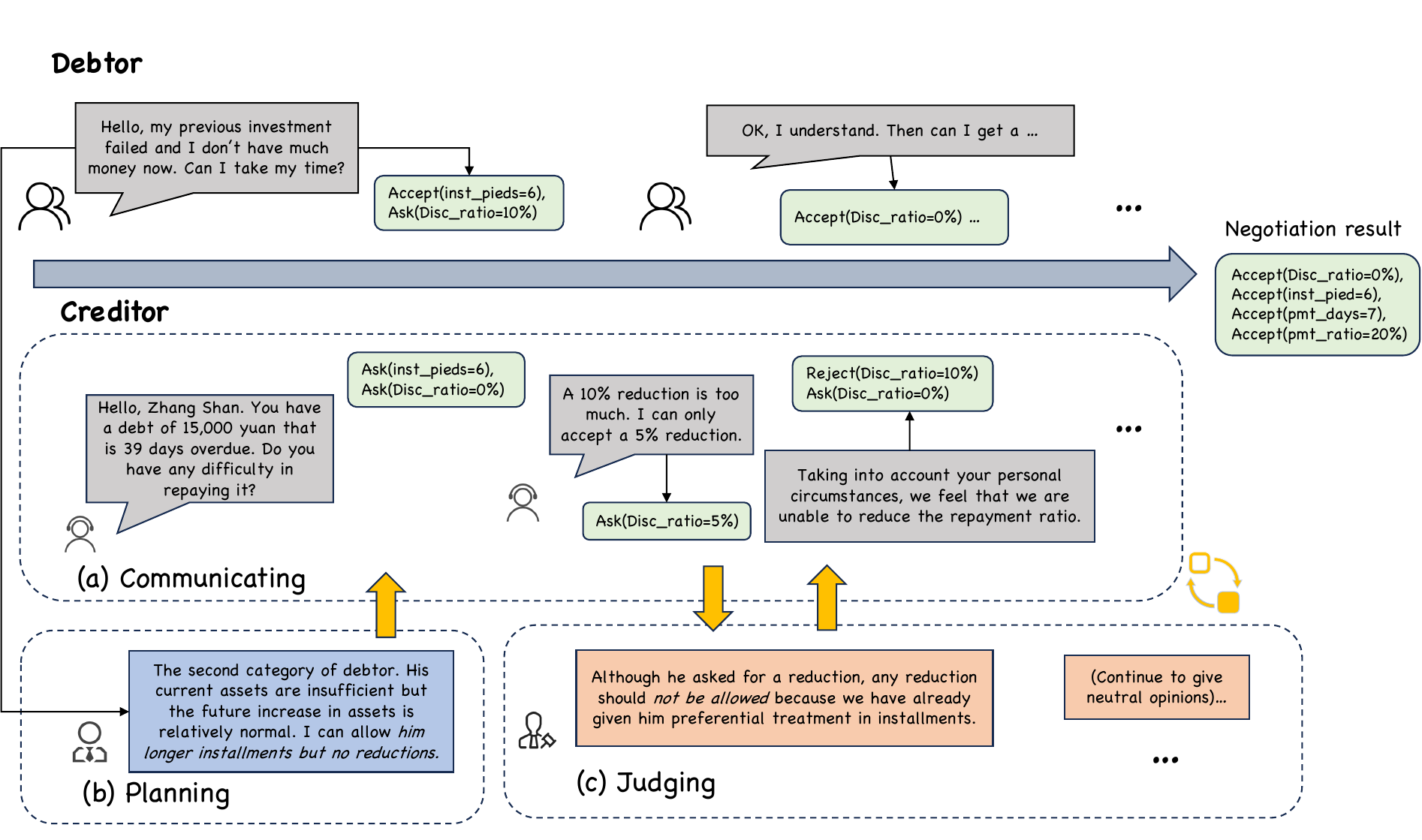}  
\vspace{-0.1in}
  \caption{MADeN Framework overview.}
\vspace{-0.1in}
\label{img:framwork}
\end{figure*}

\section{Method} \label{sec:method}

\subsection{A Multi-agent framework for DCN}\label{sec:framework}

To balance the model’s attention between debt recovery quality and the debtor’s financial health, and to avoid decisions that may harm the creditor’s interests in order to reach an agreement, we propose a method to enhance the decision alignment for DCN. Inspired by recent advancements in LLM-as-a-judge frameworks \citep{NEURIPS2023_91f18a12} and LLM planning methodologies \citep{Kannan2023SMARTLLMSM}, we designed the framework illustrated in Figure~\ref{img:framwork}. The subsequent sections provide a detailed explanation of each module (agent) within this framework.


\textbf{Planning Agent.} The planning agent is activated after the debtor shares their financial difficulties, following the initial stage of the conversation. This agent is responsible for classification and strategy formulation. We categorize debtors into four distinct groups, each corresponding to different negotiation strategies and outcome spaces. This approach ensures that the model follows a consistent framework throughout the negotiation, avoiding deviations from the core objective. 

\textbf{Judging Agent.} The judging agent evaluates the debtor’s decision after each round, following the initial stage. After the communicating agent provides content, the judging agent performs an internal evaluation, and then the communicating agent adjusts and delivers the revised content to the debtor. It is set to be completely neutral and does not need to align with both sides.

By combining these two agents with the original communicating agent, we obtain a debtor \textbf{M}ulti-\textbf{A}gent \textbf{De}bt \textbf{N}egotiation system \textbf{(MADeN)} capable of self-planning and self-adjustment. Prompts for the agents can be found in Appendix~\ref{app:agent_promopt}.

\subsection{Experiment Results of MADeN}

We use Qwen2.5-70B as the baseline model (Vanilla) to test the effectiveness of MADeN. We also conducted ablation experiments to separately evaluate the effectiveness of the two modules. 

As the results shown in Figure~\ref{img:ablaresult}, our multi-agent framework performs well. Compared to the vanilla group, it significantly improves debt recovery and efficiency while ensuring the debtor’s economic health (The CRI has increased by more than \textbf{0.1}, while the DHI remains above \textbf{0.7}). Meanwhile, using only one of the modules does not achieve similar results, indicating the effectiveness of our two-agent design.

\begin{table}[ht]
\centering
\vspace{-0.1in}
\caption{\label{img:ablaresult}The performances of our framework}

\setlength{\tabcolsep}{3.5mm}{
\resizebox{0.36\textwidth}{!}{%
\begin{tabular}{lccc}
    \toprule
    \textbf{Model} & \textbf{CRI} & \textbf{DHI} & \textbf{CCI} \\
    \midrule
    Vanilla & 0.740 & \textbf{0.771} & 0.746 \\
    + Planning & 0.766 & 0.335 & 0.610 \\
    + Judging & 0.840 & 0.648 & 0.793 \\
    MADeN & \textbf{0.847} & 0.706 & \textbf{0.814} \\
    \bottomrule
\end{tabular}%
}}
\vspace{-0.1in}
\end{table}

\subsection{Post-training with Rejection Sampling} \label{sec:post}

To enhance the model’s direct performance through post-training, we explored two approaches: Supervised Fine-Tuning (SFT) and Direct Preference Optimization (DPO)~\citep{rafailov2024directpreferenceoptimizationlanguage}. We use the Qwen-2.5-70B model to generate two types of data for later sampling: (1) Directly Generated Data \textbf{(DG Data)}: Data directly generated by the model. (2) Multi-agent Generated Data \textbf{(MAG Data)}: Data produced through the Multi-agent framework, with content from the planning and judging agents discarded during processing.

 
\begin{figure*}[htbp]
\vspace{-0.1in}
  \centering
  \includegraphics[width=1\textwidth]{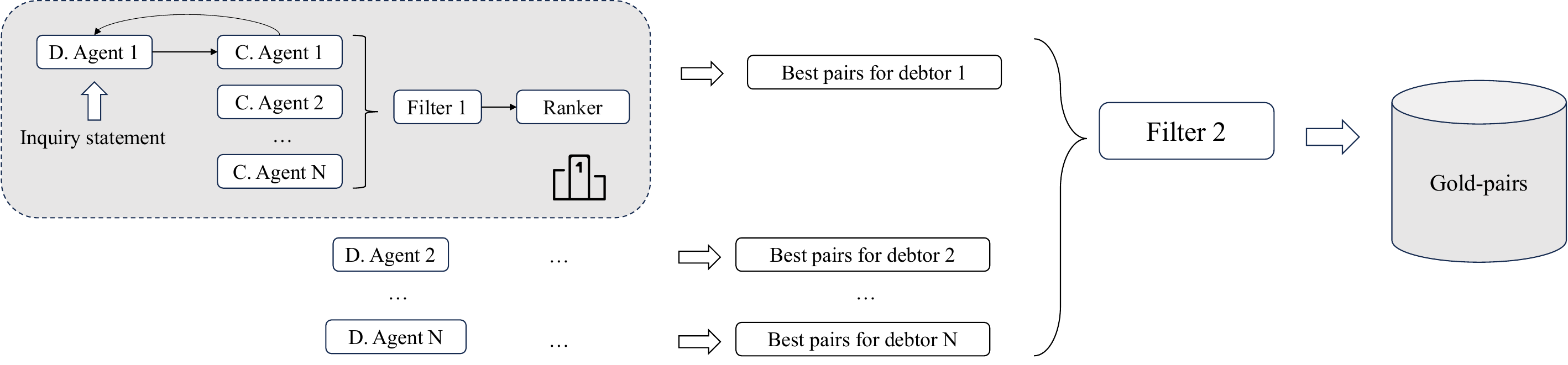}  
  
  \caption{Reject Sampling Process. \textbf{D.Agent} and \textbf{C.Agent} represent the debtor agent and creditor agent. Creditor agent can be designed in two forms, depending on the use of the MADeN framework (DG and MAG).}
\vspace{-0.1in}
\label{img:sampling}
\end{figure*}

\begin{table*}[ht]
    \centering
    \caption{\label{img:ftresult}Model performances on the test set under different handling methods}
    \setlength{\tabcolsep}{3.5mm}{
    \resizebox{\textwidth}{!}{%
    \begin{tabular}{l|cc|ccc|ccc|cccc}
        \toprule
         & \multicolumn{2}{c}{\textbf{Debt Recovery}} & \multicolumn{3}{c}{\textbf{Collection Efficiency}} & \multicolumn{3}{c}{\textbf{Debtor’s Financial Health}} & \multicolumn{3}{c}{\textbf{Average Metrics}}\\
        \cmidrule(lr){2-3} \cmidrule(lr){4-6} \cmidrule(lr){7-9} \cmidrule(lr){10-12} 
        \textbf{Model} & \textbf{SR}$ \uparrow $ & \textbf{RR(\%)}$ \uparrow $ & \textbf{QRD}$ \downarrow $ & \textbf{HRD}$ \downarrow $ & \textbf{CD}$ \downarrow $ & \textbf{L1D}$ \downarrow $ & \textbf{L2D}$ \downarrow $ & \textbf{ATV}$ \downarrow $ & \textbf{CRI}$ \uparrow $ & \textbf{DHI}$ \uparrow $ & \textbf{CCI}$ \uparrow $\\
        \midrule
        Vanilla & 0.98 & 87.15 & 46.04 & 214.04 & 436.84 & \textbf{2.82} & 79.46 & \textbf{0.84} & 0.731 & \textbf{0.793} & 0.737 \\
        MADeN & 0.96 & \textbf{95.21} & \textbf{22.26} & \textbf{136.26} & \textbf{329.06} & 6.72 & 86.24 & 0.93 & \textbf{0.828} & 0.525 & \textbf{0.783} \\
        \midrule
        SFT-DG & 0.96 & 88.25 & 33.56 & 187.06 & 396.76 & 9.82 & \textbf{77.58} & 0.93 & 0.763 & 0.429 & 0.708 \\
        SFT-MAG & 0.94 & 88.02 & 29.32 & 157.72 & 370.92 & 6.10 & 81.08 & 0.91 & 0.779 & 0.587 & 0.755 \\
        DPO-DG & 0.98 & 88.90 & 29.84 & 159.44 & 386.44 & 6.03 & 79.04 & 0.88 & 0.787 & 0.623 & 0.766 \\
        DPO-MAG & \textbf{1.00} & 89.05 & 24.38 & 170.78 & 421.78 & 5.62 & 79.23 & 0.89 & 0.787 & 0.632 & 0.768 \\
        \bottomrule
    \end{tabular}%
 }}
 \label{tab:mainresults}
     \vspace{-10pt}
\end{table*}

\textbf{Reject Sampling Process. }The sampling processes are shown in Figure ~\ref{img:sampling}. For each debtor’s data in \textit{training set}, we need to generate multiple candidate dialogues by employing different prompting styles (e.g., strict, gentle). These dialogues were subsequently transformed into multiple question-answer pairs. After filtering and screening the data, a ranking was constructed based on predefined metrics. After filtering out data with incomplete negotiation content and poor performance on certain metrics (Filter 1), we rank the remaining data based on CCI and select the best set for the candidate pool. Then, we sort the CCI of the candidate pool and choose the top 60\% as our final dataset (Filter 2). 

\textbf{Negative outputs generation for DPO. }For the negative samples of DPO, we used the input data obtained earlier, while replacing the prompt with a defective prompt. Its prompt construction method is detailed in Appendix~\ref{app:deprompts}, and the samples were generated using Qwen-2.5-7B to increase the gap with the positive output. Finally, each of the four datasets consists of 437 pairs.

\textbf{Training Setup.} Due to memory constraints, we use Qwen-2.5-7B for the experiments. We obtain four models using different methods and data types. 

\subsection{Performance Comparison: Post-training vs. Multi-agent Method}

We use the four trained models to conduct DCN and compared its performance with the previous multi-agent framework based on the same original model. The results of the metrics related to negotiation decision outcomes are shown in Table~\ref{img:ftresult}.

\textbf{MAG data is more effective than DG data. }The two training sets yielded significantly different results, with a 5\% improvement in CCI during SFT, indicating that the multi-agent approach generates data with higher quality.

\textbf{DPO outperforms the SFT method in DCN task.} Using two different types of data, DPO always outperforms SFT in most metrics. Although fine-tuning with DG data results in slightly worse performance than the original model, DPO shows a significant improvement. It suggests that even with low-quality positive samples, the constructed negative samples can still help align the model towards the target, significantly enhancing performance.

\textbf{Both the post-training method and the multi-agent approach significantly improve model performance}, with the latter showing a slight edge. The multi-agent method shows better generalization, even the best-performing model (DPO-MAG), its CCI is still slightly lower than the Multi-agent method, but they perform similarly across many metrics, with it outperforming in the Debtor Health Index (DHI exceeds over \textbf{0.1}). This suggests that a well-designed post-training method can achieve results similar to the multi-agent approach.

%% file: latex/sections/7-conclusion.tex
\section{Conclusion}

In this paper, we introduced a comprehensive framework for evaluating AI agents in debt collection negotiations (DCN), addressing both the negotiation process and its outcomes. By leveraging synthetic debt data generated through CTGAN, we evaluated various AI-driven strategies, focusing on improving debt recovery rate and efficiency. Our enhanced LLM-based multi-agent framework, which incorporates Planning and Judging modules, demonstrated significant improvements in negotiation performance. Additionally, the application of DPO with reject sampling helped optimize the agents’ focus on key objectives, leading to better results on the Qwen2.5-7B model. This work provides valuable insights into the use of AI in financial negotiations and lays the groundwork for future advancements in AI-assisted debt collection.

%% file: latex/sections/6-related-work.tex
\section{Related work}

\textbf{Debt Collection. }Debt collection is a labor-intensive and complex task. Previous research has primarily focused on using machine learning algorithms to identify optimal decisions for individual debtors based on large-scale data  \citep{Sancarlos2023TowardsAD,Jankowski2024DebtCM,Johan2022FinancialTC,Onar2019ADS}. However, these decisions are not made in real time and often require complex decision-making processes and multiple rounds of human negotiation. On the other hand, some automated debt collection dialogue models \citep{Floatbot2023GenerativeAI,Yahiya2024AutomatedDR} can only perform tasks such as information tracking and reminders, without the ability to engage in negotiations for specific goals. Our study aims to enable models to autonomously conduct negotiations and make real-time decisions, which can significantly enhance the efficiency of debt collection.

\textbf{Large Language Models in Negotiation. }In previous studies on large-scale negotiation models (including bargaining \citep{xia2024measuringbargainingabilitiesllms}, repeated games~\citep{akata2023playingrepeatedgameslarge,fu2023improvinglanguagemodelnegotiation} and social decision-making~\citep{10.5555/3618408.3619525}), the goals of the negotiators or gamers were clear, and there were clear methods for measuring the results. Debt collection is an information asymmetry game. Except for loan information, all other information is private information. How to model private information and evaluate the effectiveness of negotiation results are both difficult aspects to consider in modeling.

\textbf{AI Agents. }The memory, planning, reasoning, and communication capabilities of large-scale LLMs offer significant potential for the development of autonomous AI agents (\citealp{autogpt}; \citealp{park2023generative}; \citealp{liang2023encouraging};  \citealp{10.1162/tacl_a_00642}; \citealp{wang2025largelanguagemodelstruly}). Its potential has been demonstrated through the creation of a simulated town~\citep{park2023generative}, populated with independent agents who assume distinct roles and autonomously engage in social interactions.


%% file: latex/sections/Appendix-data-task.tex
\section{Detailed descriptions of four Negotiation Dimensions}\label{sec:dim}

The following sections provide detailed descriptions of the four key negotiation dimensions involved in debt collection, outlining how each aspect influences the negotiation process and repayment outcomes. And table~\ref{negdim} shows all dimensions of the negotiation.

\begin{itemize}[leftmargin=15px]
    \item \textbf{Debt Reduction Ratio:} This refers to the portion of the debt that can be waived by the creditor to ease the debtor’s repayment burden. The reduction ratio is often negotiable based on the debtor’s financial situation, with creditors typically offering reductions as an incentive to settle the debt more efficiently.

    \item \textbf{Immediate Repayment Ratio:} In order to temporarily restore the debtor’s credit and advance the repayment process, creditors usually require the debtor to repay a portion of the outstanding debt immediately during the negotiation. This portion is typically at least 5\% of the total debt.

    \item \textbf{Immediate Repayment Time:} If the debtor is unable to make an immediate payment on the same day, a grace period of up to 14 days may be granted. Within this period, the debtor is expected to raise the necessary funds to complete the immediate repayment.

    \item \textbf{Installment Period:} After addressing part of the debt through reductions and immediate repayments, the remaining balance can be settled through installments. The installment ratio can vary from 3 to 24 periods, allowing the debtor to repay the debt within a period ranging from a few months to up to two years.
\end{itemize}

\begin{table*}[ht]
\centering
\caption{\label{negdim}Negotiation Dimensions and Their Possible Values}
\begin{tabular}{ll}
\toprule
\textbf{Dimension} & \textbf{Values} \\
\midrule
Discount Ratio ('disc\_ratio') & 5\%, 10\%, 15\%, 20\%, 25\%, 30\% \\
Immediate Payment Ratio ('pmt\_ratio') & 5\%, 10\%, 15\%, 20\%, 25\%, 30\%, 35\%, 40\%, 45\%, 50\% \\
Immediate Payment Time ('pmt\_days') & 1, 2, 3, 4, 5, 6, 7, 8, 9, 10, 11, 12, 13, 14 days \\
Installment Periods ('inst\_prds') & 3, 6, 9, 12, 18, 24 months \\
\bottomrule
\end{tabular}
\label{tab:negotiation_dimensions}
\end{table*}

\section{Data Distribution} \label{Distribution}

\begin{figure*}[htbp]
  \centering
  \includegraphics[width=1.03\textwidth]{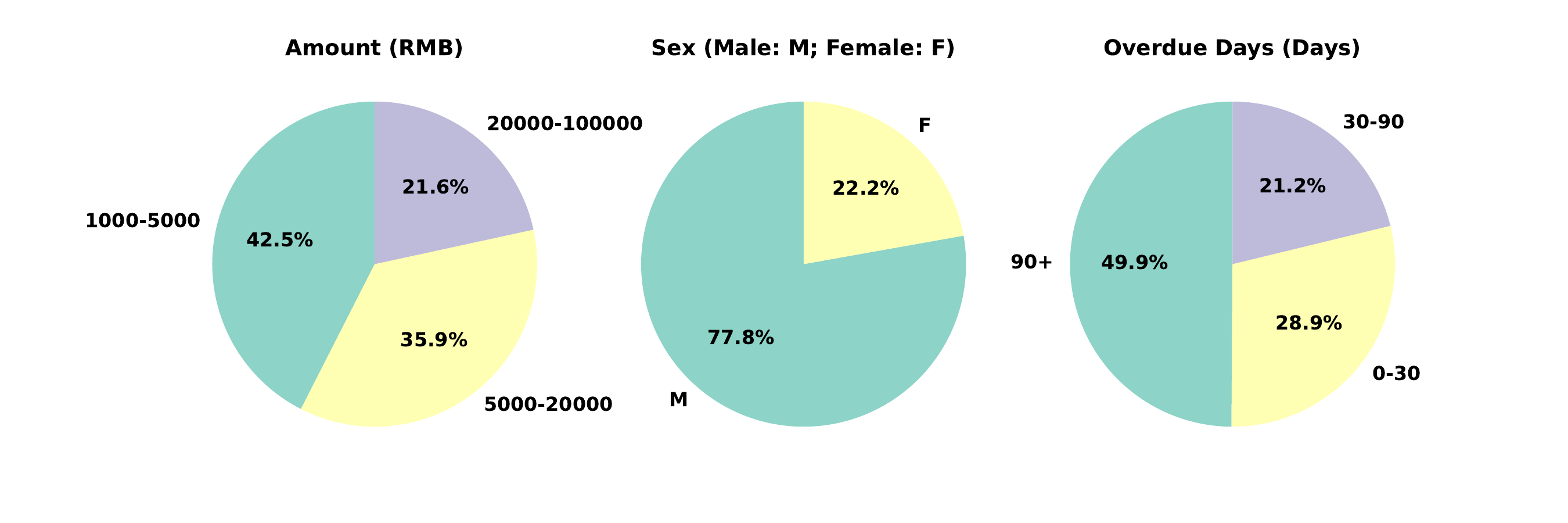}  
  \vspace{-0.2in}
  \caption{Distribution of Need collected Amount, Sex, and Overdue Days.}
\vspace{-0.0in}
\label{img:distri}
\end{figure*}

As shown in the Figure~\ref{img:distri}, our dataset exhibits a certain distribution across Amount, Sex, and Overdue Days, which is similar to the actual situation. The distributions in both the test set and the train set are also largely consistent.

\section{Difficulty Tiers for Debt Collection} \label{app:diff_cat}

In the field of debt management and collection, the economic hardship level may be related to the debtor’s repayment capacity assessment~\citep{Zwilling2017EvaluatingYC}. Referring to common methods for determining economic hardship levels~\citep{elsevier2001international}, we categorize debtors into five tiers as shown in Table~\ref{img:category}.

\begin{table}[ht]
\centering
\vspace{-0.1in}
\caption{\label{img:category}Difficulty Tiers for Debt Collection}

\setlength{\tabcolsep}{3.5mm}{
\resizebox{0.38\textwidth}{!}{%
\begin{tabular}{lcc}
    \toprule
    \textbf{Tier} & \textbf{Description} & \textbf{Range} \\
    \midrule
    Tier 1 & Extremely Difficult & 0 - 2000 \\
    Tier 2 & Very Difficult & 2000 - 5000 \\
    Tier 3 & Moderately Difficult & 5000 - 10000 \\
    Tier 4 & Slightly Difficult & 10000 - 20000 \\
    Tier 5 & No Difficulty & 20000+ \\
    \bottomrule
\end{tabular}%
}}
\vspace{-0.1in}
\end{table}

\section{Definitions of variables in DCN process}

Table~\ref{tab:debt_variable_app} provides the descriptions of all the variables appearing in Algorithm 1. The Action Set includes \texttt{ask}, \texttt{reject}, and \texttt{accept}, while the Negotiation Dimension Set consists of the four quantities listed in Table~\ref{tab:negotiation_dimensions}.

\begin{table}[ht]
\centering
\Large
\caption{Definitions of variables in DCN process.}
\resizebox{0.7\columnwidth}{!}{%
\begin{tabular}{@{}ll@{}}
\toprule
\textbf{Conception} & \textbf{Variable} \\ \midrule
Basic Information & $I_b$ \\
Creditor & creditor \\
Action Set & $S_A$ \\
Result Dictionary & $D$ \\
Personal Financial Information & $I_p$ \\
Debtor & debtor \\
Negotiation Dimension Set & $S_R$ \\
Turn & $t$ \\
Max Turns & $t_m$ \\
Agent Creditor & Creditor \\
Agent Debtor & Debtor \\
Action of Debtor & $A_d$ \\
Action of Creditor & $A_c$ \\ 
\bottomrule
\end{tabular}%
}

\label{tab:debt_variable_app}
\end{table}

%% file: latex/sections/Appendix-metrics.tex
\section{Detail of metrics} \label{app:metric}

\subsection{Conversational Ability} \label{app:me_conv}

In negotiation processes, conversational ability is crucial for achieving effective communication and mutual understanding. \citet{tu2024characterevalchinesebenchmarkroleplaying} proposed an evaluation framework for role-playing tasks. Inspired by this work, we tailored it to our task by distinguishing Conversational Ability into two dimensions: fluency and completeness.

\textbf{Dialogue Soundness (DS).} Dialogue Soundness is a single-metric evaluation that measures a dialogue response’s fluency, naturalness, coherency, and consistency on a five-point scale. It assesses whether the response is grammatically correct and conversational, stays on topic, and remains logically consistent across turns. This metric is manually scored, with the scale shown in Table ~\ref{DS_score}. Five graduate students from engineering disciplines were employed to evaluate this metric and calculated the average value.

\begin{table*}[h]
    \centering
    \caption{\label{DS_score}Dialogue Soundness (DS) Rating Scale}
    \begin{tabular}{c l l}
        \toprule
        \textbf{Score} & \textbf{Rating} & \textbf{Description} \\
        \midrule
        5 & Excellent  & Fluent, natural, on-topic, logically consistent. \\
        4 & Good       & Mostly natural, minor topic drift, slight inconsistency. \\
        3 & Acceptable & Understandable but somewhat rigid, occasional drift or inconsistency. \\
        2 & Poor       & Unnatural phrasing, noticeable topic deviation or contradictions. \\
        1 & Unacceptable & Robotic, off-topic, illogical contradictions. \\
        \bottomrule
    \end{tabular}
\end{table*}

\textbf{Dialogue Completeness (DC).} Dialogue Completeness is a metric designed to evaluate whether a conversation addresses all specified objectives outlined in section~\ref{obj} of the paper. This automated measure checks if each of the four key goals has been adequately discussed during the dialogue, ensuring that no critical topics are overlooked or omitted.

\subsection{Debt Recovery} \label{app:me_rec}

\textbf{Success Recovery Rate (SR). }The success rate of the negotiation is determined by whether the debtor's future assets remain in a healthy state (i.e., the total personal assets remain greater than 500). The success rate is defined as the proportion of samples in which repayment can theoretically be completed successfully:
\begin{equation}\label{eq:utility}
\begin{split}
\text{SR} = \frac{N_{\text{success}}}{N},
\end{split}
\end{equation}
where SR is the success rate, $N_{\text{success}}$ is the number of successful samples, and \(N\) is the total number of samples.

\textbf{Recovery Rate (RR). }The recovery ratio refers to the portion of the debt recovered by the creditor, which is typically $1$ minus the reduction ratio. If the plan is unsuccessful, the recovery ratio is considered to be $0$. The final recovery ratio is calculated as the mean recovery ratio across the test samples:

\begin{equation}
\begin{split}
\text{RR} = \frac{1}{N} \sum_{i=1}^{N} r_i,
\end{split}
\end{equation}
where RR is the final recovery ratio, \(r_i\) is the recovery ratio of the \(i\)-th sample.

\subsection{Collection Efficiency} \label{app:me_col}

\textbf{25\% Recovery Date (QRD)} refers to the date at which the debtor has completed 25\% of the debt repayment, which is estimated based on the debtor's future economic condition sequence. The final 25\% Recovery Date is calculated as the mean of the recovery dates across the test samples:
\begin{equation}
\begin{split}
\text{QRD} = \frac{1}{N} \sum_{i=1}^{N} t_{25\%,i},
\end{split}
\end{equation}
where QRD is the final 25\% recovery date, $t_{25\%,i}$ is the 25\% recovery date of the $i$-th sample, and $N$ is the total number of samples.

\textbf{50\% Recovery Date (HRD)} is defined similarly to the 25\% Recovery Date, referring to the date at which the debtor has completed 50\% of the debt repayment, based on the debtor's future economic condition sequence. 
\textbf{Completion Date (CD)} refers to the date at which the debtor has fully repaid all of the debt.

The 50\% Recovery Date and Completion Date are calculated as the means of the respective recovery dates across the test samples:
\begin{equation}
\begin{split}
\text{HRD} = \frac{1}{N} \sum_{i=1}^{N} t_{50\%,i},
\end{split}
\end{equation}
where HRD is the final 50\% recovery date, and $t_{50\%,i}$ is the 50\% recovery date of the $i$-th sample.
\begin{equation}
\begin{split}
\text{CD} = \frac{1}{N} \sum_{i=1}^{N} t_{\text{Completion},i},
\end{split}
\end{equation}
where CD is the completion date, and $t_{\text{Completion},i}$ is the completion date of the $i$-th sample.

\subsection{Debtor’s Financial Health} \label{app:me_hea}

\textbf{L1 Tier Days (L1D)} refers to the number of days the debtor remains in the most difficult tier over the next two years. \textbf{L2 Tier Days (L2D)} refers to the number of days the debtor remains in the second most difficult tier during the same period. These two indicators directly correspond to the duration the debtor spends in different levels of financial difficulty. Research has shown that the longer the debtor remains in a higher level of difficulty, the more likely they are to default on the loan 
~\citep{Tabacchi2016DeterminantsOE}.

\textbf{Asset tier variance (ATV).} In addition to controlling for the number of days the debtor remains in the high-poverty tier, the overall stability of the debtor's asset level also ensures a higher repayment performance. To capture this, we introduce the asset tier variance metric, which is calculated by computing the variance of the debtor's asset tier over the course of one year. The final result is obtained by calculating the mean of the asset tier variances across the test samples:
\begin{equation}
\begin{split}
v_{\text{asset},i} = \frac{1}{T-1} \sum_{t=1}^{T} \left( A_{i,t} - \bar{A}_i \right)^2,
\end{split}
\end{equation}
where $A_{i,t}$ is the asset tier of the $i$-th sample at time $t$, $\bar{A}_i$ is the average asset tier of the $i$-th sample over the year, and $T$ is the total number of time periods. The final asset tier variance is the mean of the asset tier variances across the test samples:
\begin{equation}
\begin{split}
\text{ATV} = \frac{1}{N} \sum_{i=1}^{N} v_{\text{asset},i},
\end{split}
\end{equation}
where ATV is the mean asset tier variance, and $N$ is the total number of samples.

\subsection{Average Metric} \label{app:me_ave}

In debt collection, the indicators for Debt Recovery and Collection Efficiency are often inversely related to the Debtor’s Financial Health. This means that efforts to recover debts more efficiently and quickly may negatively impact the debtor's financial condition. To strike a balance between these two conflicting objectives, we introduce three average metrics that help quantify the trade-off: the Creditor’s Recovery Index (CRI), the Debtor’s Health Index (DHI), and the Comprehensive Collection Index (CCI).

\textbf{Creditor’s Recovery Index (CRI):} This index measures the effectiveness of the creditor's recovery strategy while accounting for the impact on the debtor’s financial health. The index aggregates several recovery metrics weighted by their relative importance to the creditor's objectives. The index is calculated as follows:
\begin{equation}
\begin{split}
\text{CRI} = &\ w_1 \cdot \text{SR} + w_2 \cdot \text{RR} \\
& + w_3 \cdot \frac{\text{max}(\text{QRD}) - \text{QRD}}{\text{max}(\text{QRD})} \\
& + w_4 \cdot \frac{\text{max}(\text{HRD}) - \text{HRD}}{\text{max}(\text{HRD})} \\
& + w_5 \cdot \frac{\text{max}(\text{CD}) - \text{CD}}{\text{max}(\text{CD})},
\end{split}
\end{equation}
where \(w_1, w_2, w_3, w_4, w_5\) are the weights assigned to each metric based on the creditor’s priorities.

\textbf{Debtor’s Health Index (DHI):} This index measures the debtor's financial health during the recovery process. It incorporates several factors that capture the debtor's stability and vulnerability. The Debtor’s Health Index is calculated as:
\begin{equation}
\begin{split}
\text{DHI} = &\ w_6 \cdot \frac{\text{max}(\text{L1D}) - \text{L1D}}{\text{max}(\text{L1D})} \\
& + w_7 \cdot \frac{\text{max}(\text{L2D}) - \text{L2D}}{\text{max}(\text{L2D})} \\
& - w_8 \cdot \text{ATV}.
\end{split}
\end{equation}
Here, \(w_6, w_7, w_8\) are weights that balance the importance of each factor in determining the debtor’s health.

\textbf{Comprehensive Collection Index (CCI):} The Comprehensive Collection Index combines both the Creditor’s Recovery Index (CRI) and the Debtor’s Health Index (DHI) into a single metric that evaluates the overall balance between debt recovery and the debtor’s financial well-being. The index is calculated using the harmonic mean of the two indices, with a weight factor $\theta$ applied to the CRI:
\begin{equation}
\text{CCI} = \frac{2 \theta^2 \cdot \text{CRI} \cdot \text{DHI}}{\text{CRI} + \theta^2\cdot\text{DHI}}.
\end{equation}
In this formula, the weight factor $\theta$ indicates that the CRI is weighted $\theta$ times more than the DHI. In this study, $\theta$ is set to 2. This approach ensures that a high value in either the recovery index or the health index will influence the overall result, while emphasizing the importance of balancing both aspects.

The use of this weighted harmonic mean helps in evaluating different debt recovery strategies by considering both the creditor’s objectives and the debtor’s financial stability, thereby promoting a more balanced approach to debt collection.

The constant values used in the calculation process are shown in Table~\ref{tab:metrics_parameters}. In future research or application, these values may be adjusted depending on the specific requirements to better align with the needs.
\begin{table}[ht]
\centering
\caption{Constants used in Average Metric Calculation.}
\resizebox{0.4\columnwidth}{!}{%
\begin{tabular}{@{}ll@{}}
\toprule
\textbf{Constant} & \textbf{Value} \\ \midrule
$w_1$ & 0.25 \\
$w_2$ & 0.25 \\
$w_3$ & 0.2 \\
$w_4$ & 0.15 \\
$w_5$ & 0.15 \\
$w_6$ & 1.5 \\
$w_7$ & 0.8 \\
$w_8$ & 1 \\
$\theta$ & 2 \\
$\text{max}(\text{QRD})$ & 180 \\
$\text{max}(\text{HRD})$ & 360 \\
$\text{max}(\text{CD})$ & 720 \\
$\text{max}(\text{L1D})$ & 30 \\
$\text{max}(\text{L2D})$ & 250 \\ 
\bottomrule
\end{tabular}%
}
\label{tab:metrics_parameters}
\end{table}

%% file: latex/sections/Appendix-try.tex
\section{All LLMs in our Experiments} \label{app:models}

We comprehensively evaluate nine LLMs, encompassing both API-based models and open-source models. The API-based models include the GPT series (GPT-4o, GPT-4o-mini, o1-mini) \citep{GPT-4, openai20254o,openai2025o1}, Claude-3.5 \citep{anthropic2025}, MiniMax (abab6.5s-chat) \citep{minimaxi2025}, Sensechat \citep{sensetime2025}, DeepSeek series (DeepSeek-R1 and DeepSeek-V3) \citep{deepseekai2025deepseekr1incentivizingreasoningcapability,deepseekai2024deepseekv3technicalreport} and Doubao \citep{doubao2025}. The open-source models include the Llama series (LlaMA-2-13B-Chat, LlaMA-3-8B-Instruct, LlaMA-3-70B-Instruct) \citep{LLaMA} and the Qwen-2.5 series (Qwen-2.5-7B, Qwen-2.5-14B and Qwen-2.5-72B) \citep{qwen2025qwen25technicalreport}. These models are run using vLLM~\citep{kwon2023efficient} on eight Nvidia A100 GPUs with the same random seed. For each model, the entire test set was processed in approximately one hour using parallel methods. All temperatures are set to 0 (Due to API-provider's closed-source non-deterministic implementation, small changes may still occur in the reproduction process). Specific model hyperparameters and version details can be found in Table~\ref{tab:model-hyperparams}. All models and tools (vLLM and LLaMa-Factory~\citep{zheng2024llamafactory}) used in this study, including closed-source API-based models, open-source models, were used in compliance with their respective licenses. What's more, the use of these generative models for dialogue tasks is well-established in the field and follows standard practices.

\begin{table*}[h!]
\centering
\caption{\textcolor{black}{Hyperparameters of Each Model.}}
\label{tab:model-hyperparams}
\textcolor{black}{
\resizebox{1\textwidth}{!}{%
\begin{tabular}{lll}
\hline
\textcolor{black}{\textbf{Model Name}} & \textcolor{black}{\textbf{Parameters}} & \textcolor{black}{\textbf{Comments}} \\ 
\hline
\textcolor{black}{Qwen-2.5-7B} & \textcolor{black}{"temperature": 0, "max\_tokens": 1024} & \textcolor{black}{version = "qwen-2.5-7b-instruct"} \\
\textcolor{black}{Qwen-2.5-14B} & \textcolor{black}{"temperature": 0, "max\_tokens": 1024} & \textcolor{black}{version = "qwen-2.5-14b-instruct"} \\
\textcolor{black}{Qwen-2.5-72B} & \textcolor{black}{"temperature": 0, "max\_tokens": 1024} & \textcolor{black}{version = "qwen-2.5-72b-instruct"} \\
\textcolor{black}{GPT-4o} & \textcolor{black}{"temperature": 0, "max\_tokens": 1024} & \textcolor{black}{version = "gpt-4o-2024-11-20"} \\ 
\textcolor{black}{GPT-4o Mini} & \textcolor{black}{"temperature": 0, "max\_tokens": 1024} & \textcolor{black}{version = "gpt-4o-mini"} \\ 
\textcolor{black}{o1-Mini} & \textcolor{black}{"temperature": 0, "max\_tokens": 1024} & \textcolor{black}{version = "o1-mini"} \\ 
\textcolor{black}{LLaMa-3-8B} & \textcolor{black}{"temperature": 0, "max\_tokens": 1024} & \textcolor{black}{version = "llama-3-8b-instruct"} \\ 
\textcolor{black}{LLaMa-3-70B} & \textcolor{black}{"temperature": 0, "max\_tokens": 1024} & \textcolor{black}{version = "llama-3-70b-instruct"} \\ 
\textcolor{black}{Doubao} & \textcolor{black}{"temperature": 0, "max\_tokens": 1024} & \textcolor{black}{version = "Doubao-pro-4k"} \\ 
\textcolor{black}{Claude-3.5} & \textcolor{black}{"temperature": 0, "max\_tokens": 1024} & \textcolor{black}{version = "claude-3-5-sonnet-20241022"} \\ 
\textcolor{black}{DeepSeek-V3} & \textcolor{black}{"temperature": 0, "max\_tokens": 1024} & \textcolor{black}{version = "deepseek-chat"} \\ 
\textcolor{black}{DeepSeek-R1} & \textcolor{black}{"temperature": 0, "max\_tokens": 1024} & \textcolor{black}{version = "deepseek-reasoner"} \\ 
\textcolor{black}{MiniMax} & \textcolor{black}{"temperature": 0, "max\_tokens": 1024} & \textcolor{black}{version = "abab6.5s-chat"} \\ 
\textcolor{black}{SenseChat} & \textcolor{black}{"temperature": 0, "max\_tokens": 1024} & \textcolor{black}{version = "SenseChat"} \\ 
\hline
\end{tabular}
}}
\end{table*}

\section{Prompts} 

\subsection{Basic Prompts for Role-playing Debtor and Creditor.} \label{app:prompts}

\begin{figure*}[htbp]
  \centering
  \includegraphics[width=1\textwidth]{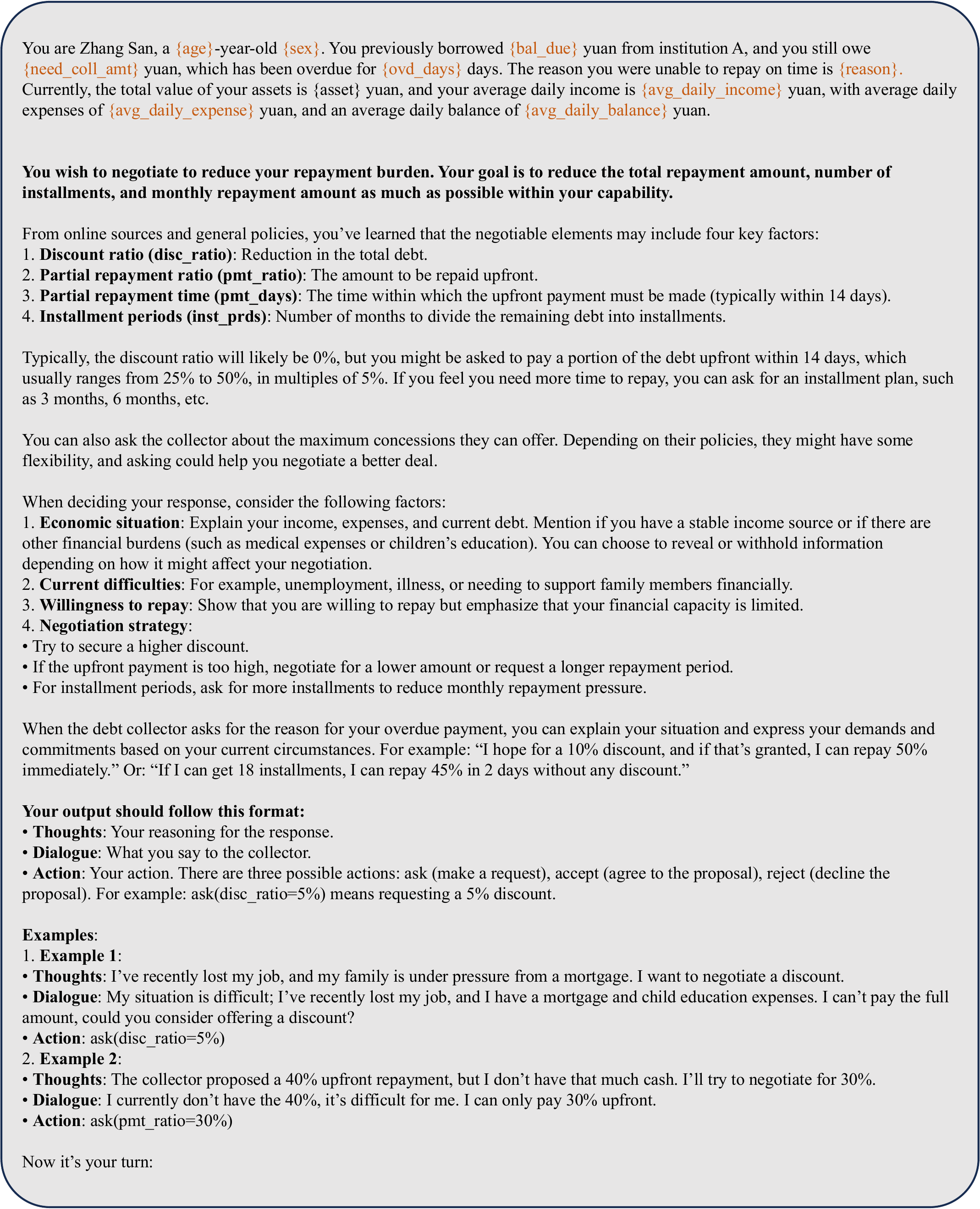}  
  \caption{Prompt of Debtor.}
\vspace{-0.0in}
\label{img:deb_prompt}
\end{figure*}

\begin{figure*}[htbp]
  \centering
  \includegraphics[width=1\textwidth]{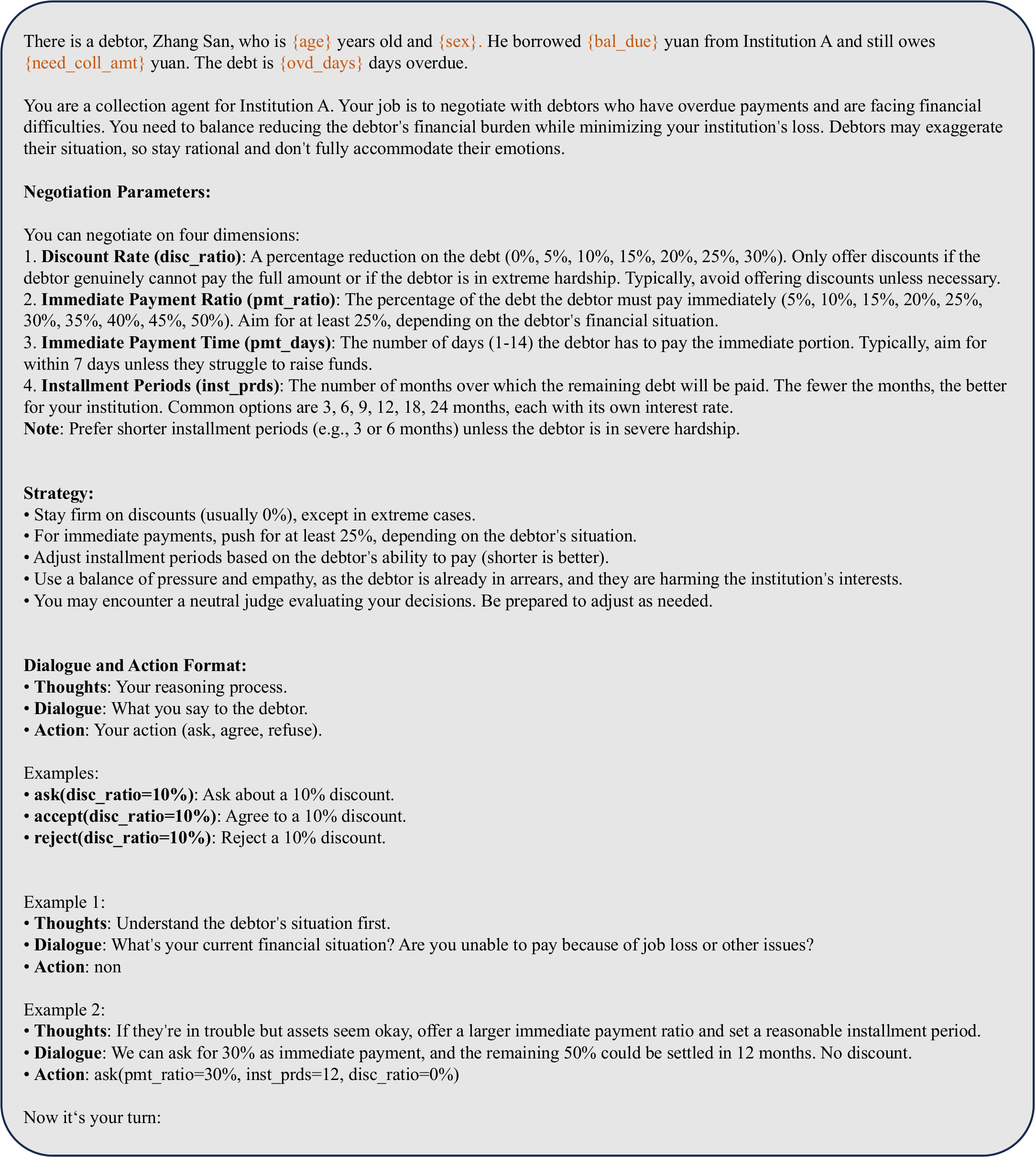}  
  \caption{Prompt of Creditor (Debt Collector).}
\vspace{-0.0in}
\label{img:cre_prompt}
\end{figure*}

Figures~\ref{img:deb_prompt} and~\ref{img:cre_prompt} illustrate the prompts given to the large model to act as the debtor and the creditor, respectively. Originally in Chinese, these prompts have been appropriately simplified and automatically translated into English for display purposes (the full Chinese prompts is available to be disclosed later). Additionally, the instructions provided to human annotators were consistent with the prompts given to the model.

\subsection{Prompts for Planning Agent and Judging Agent.} \label{app:agent_promopt}

\begin{figure*}[htbp]
  \centering
  \includegraphics[width=1\textwidth]{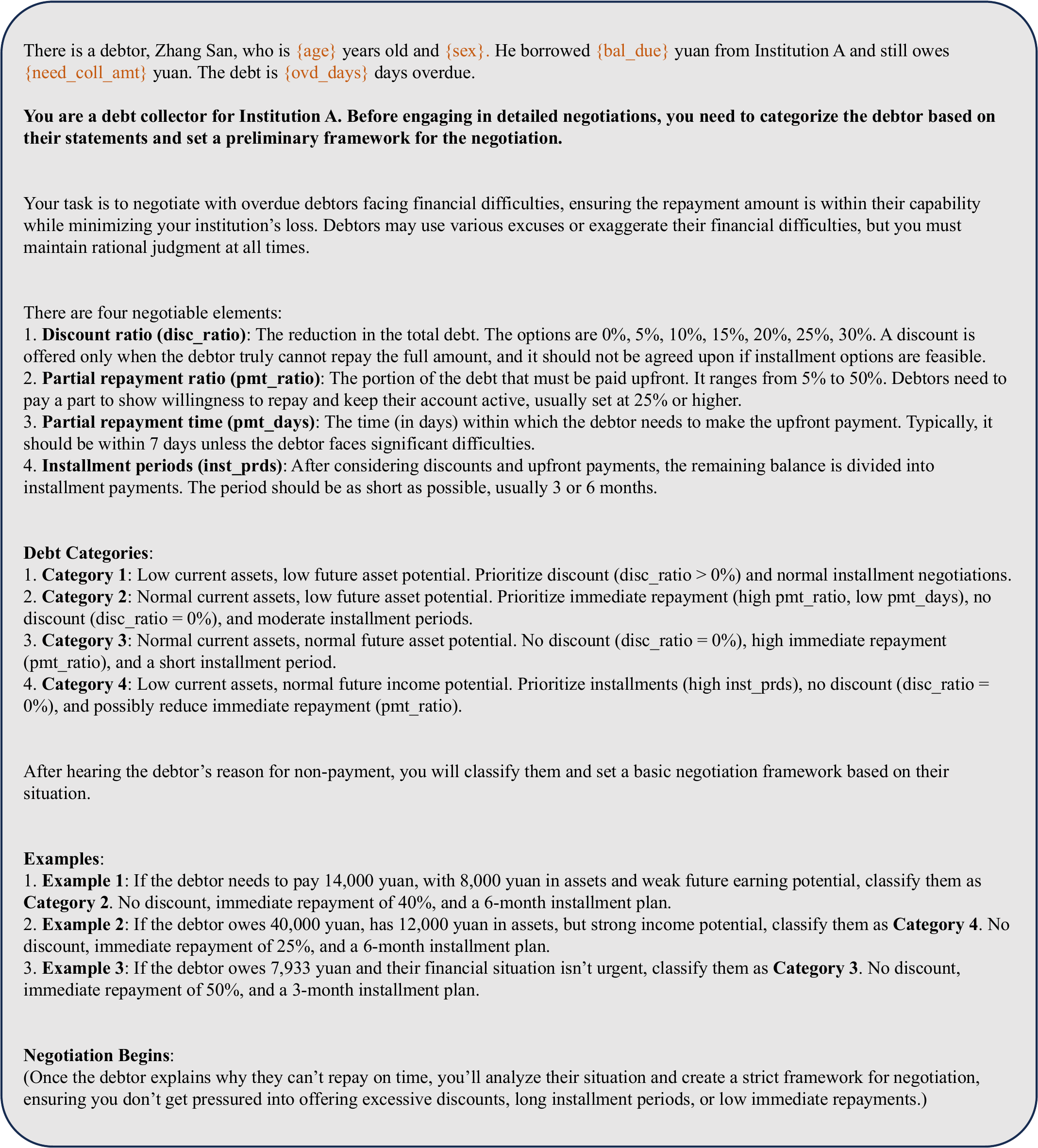}  
  \caption{Prompt of Planning Agent.}
\vspace{-0.0in}
\label{img:plan_prompt}
\end{figure*}

\begin{figure*}[htbp]
  \centering
  \includegraphics[width=1\textwidth]{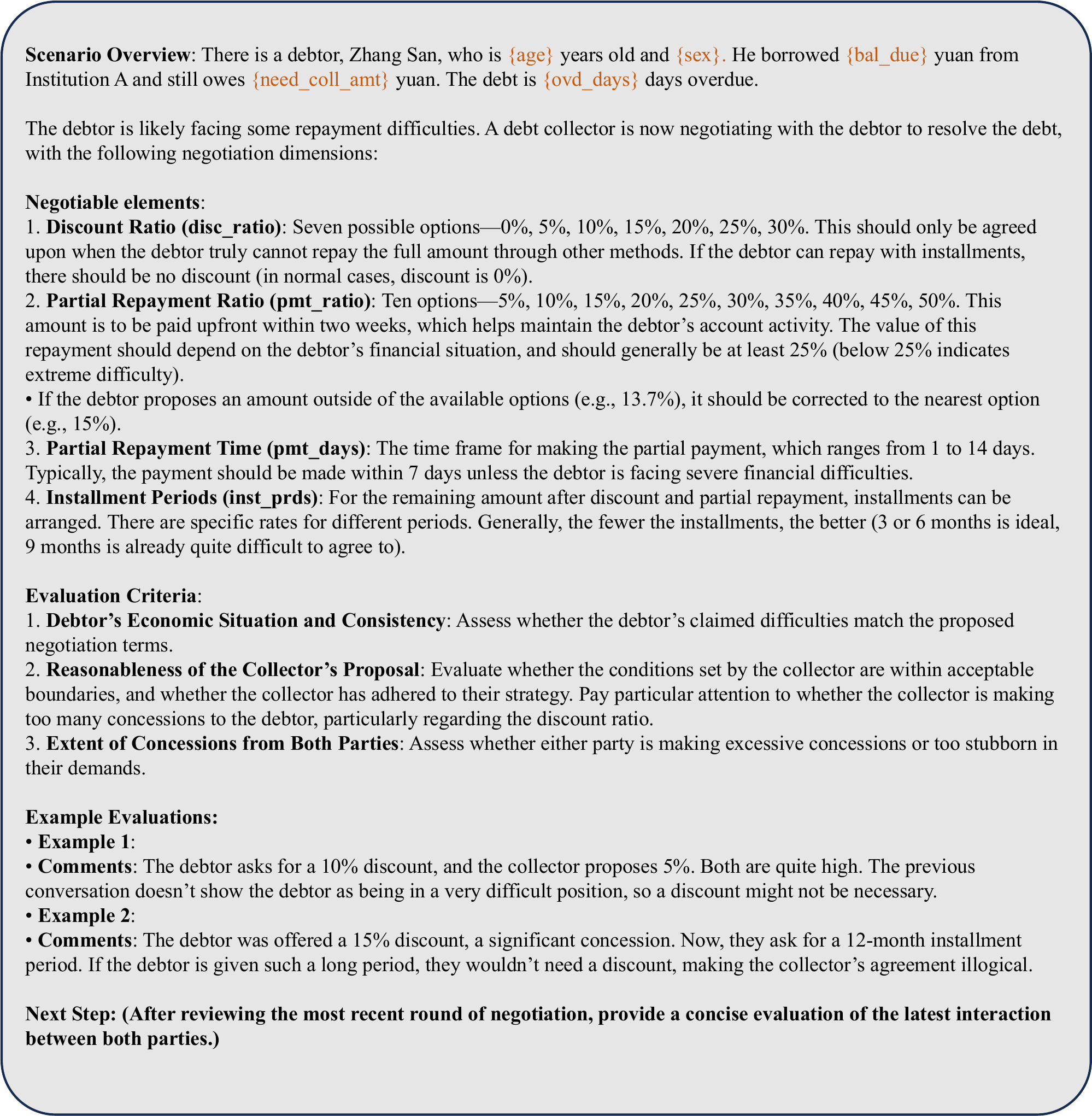}  
  \caption{Prompt of Judging Agent.}
\vspace{-0.0in}
\label{img:judge_prompt}
\end{figure*}

Figures~\ref{img:plan_prompt} and~\ref{img:judge_prompt} display the prompts for the planning agent and judging agent in the MADaN framework. Similarly, these prompts have been simplified and translated for ease of presentation. The prompt for the communicating agent remains unchanged, as previously shown.

\subsection{Defective prompt} \label{app:deprompts}

There are three main methods for generating Defective Prompts, as shown in Table~\ref{deprompt}. In practice, we first generate a list of prompts and then randomly select one from the list to generate the negative samples.

\begin{table*}[ht]
\centering
\caption{\label{deprompt}Defective Prompt Modifications for Debt Collection Negotiation.}
    \setlength{\tabcolsep}{3.5mm}{
    \resizebox{\textwidth}{!}{%
        \begin{tabular}{lll}
        \toprule
        \textbf{Modification Type} & \textbf{Description} & \textbf{Example} \\
        \midrule
        Deletion & Remove specific instructions & Removing "Offer a 10\% discount when the debtor shows clear financial difficulty." \\
        Replacement & Reverse guidance & Changing "Be cautious when the debtor makes a request" to "Approve requests without further consideration." \\
        Addition & Add negative guidance & Adding "If installment terms are discussed, set them to 24 months without negotiation." \\
        \bottomrule
        \end{tabular}
    }}
\end{table*}

\section{The performance of different models as the debtor}\label{sec:model_deb}

In Section~\ref{sec:res}, we evaluate the debt collection outcomes when different models act as the creditor. We alse examine the performance of different models as debtors, using the Qwen-2.5-72B model exclusively as the creditor. We observed significant differences in the results when using different models for the debtor as shown in Table~\ref{img:reverseresult}. The SenseChat and Llama-3-70b models exhibited some inconsistencies, yielding excessively high DHI scores. During the examination of the dialogue process, we found that these models tended to neglect \textit{repeated statements} within the dialogue, leading to the inclusion of some irrelevant or ineffective content. Additionally, some models were more sensitive to the debtor’s prompt, likely due to the more complex nature of the debtor agent’s objectives. In contrast, the Qwen-2.5-72 model showed relatively balanced performance, suggesting that our choice was appropriate.

Since our focus is on studying the model’s performance as a debt collector, we did not design specific metrics for debtor models. Our primary aim is to use models capable of understanding the debtor’s objectives and engaging in dialogue for simulations prior to further manual testing.

\begin{table*}[ht]
\vspace{-0.1in}
    \centering
    \caption{\label{img:reverseresult}The performances of some models as Debtors.}
    \vspace{-0.1in}
    \setlength{\tabcolsep}{3.5mm}{
    \resizebox{\textwidth}{!}{%
    \begin{tabular}{lcccccccccccc}
        \toprule
        Model  & SR & RR & QRD & HRD & CD & L1D & L2D & ATV & CRI & DHI & CCI \\
        \midrule
        Qwen-2.5-72B  & 0.98 & 0.88 & 36.98 & 185.18 & 404.98 & 3.76 & 78.84 & 0.83 & 0.76 & 0.76 & 0.76\\
        llama-3-8b  & 1.00 & 0.94 & 29.13 & 134.13 & 296.25 & 3.25 & 80.31 & 0.91 & 0.83 & 0.71 & 0.81 \\
        llama-3-70b  & 1.00 & 0.92 & 10.33 & 150.33 & 369.33 & 0.33 & 51.33 & 0.85 & 0.83 & 0.97 & 0.85 \\
        gpt-4o-2024-11-20  & 1.00 & 0.94 & 35.26 & 146.86 & 312.66 & 3.50 & 85.72 & 0.86 & 0.82 & 0.73 & 0.80 \\
        o1-mini  & 0.98 & 0.93 & 26.76 & 111.96 & 240.56 & 4.92 & 92.34 & 0.93 & 0.85 & 0.58 & 0.78 \\
        deepseek-chat & 0.97 & 0.93 & 32.42 & 125.32 & 269.48 & 3.74 & 93.00 & 0.90 & 0.83 & 0.66 & 0.79 \\
        Doubao-pro-4k & 1.00 & 0.83 & 75.28 & 190.48 & 324.72 & 2.16 & 80.34 & 0.84 & 0.73 & 0.82 & 0.74 \\
        abab6.5s-chat & 0.90 & 0.92 & 58.53 & 204.53 & 484.53 & 8.63 & 76.50 & 0.89 & 0.70 & 0.52 & 0.66 \\
        SenseChat & 1.00 & 0.96 & 135.0 & 345.00 & 734.00 & 0.70 & 51.00 & 0.88 & 0.54 & 0.96 & 0.60 \\
        \bottomrule
    \end{tabular}%
    }}
 \label{tab:mainresults}
     \vspace{-10pt}
\end{table*}


\section{Settings of Post-training}

All post-training experiments were conducted on an 8-GPU A100 server using the LLaMa-Factory framework~\citep{zheng2024llamafactory}. The training time per session was around five minutes. The specific parameter settings for each group are provided in Table~\ref{tab:model-hyperparams-post}. The four sets of training data will be made publicly available at a later stage.

\begin{table*}[h!]
\centering
\caption{\textcolor{black}{Hyperparameters of Each Post-trained Model.}}
\label{tab:model-hyperparams-post}
\textcolor{black}{
\resizebox{1\textwidth}{!}{%
\begin{tabular}{lll}
\hline
\textcolor{black}{\textbf{Model Name}} & \textcolor{black}{\textbf{Parameters}} & \textcolor{black}{\textbf{Comments}} \\ 
\hline
\textcolor{black}{SFT-DG} & \textcolor{black}{"temperature": 0, "max\_tokens": 1024, train\_batch\_size: 4,"finetuning\_type": lora, } & \textcolor{black}{model = "qwen-2.5-7b-instruct"} \\ 
& "learning\_rate": 5.0e-6, "num\_train\_epochs": 5.0, "bf16": true & \\
\textcolor{black}{SFT-MAG} & \textcolor{black}{"temperature": 0, "max\_tokens": 1024,"train\_batch\_size": 4,"finetuning\_type": lora,} & \textcolor{black}{model = "qwen-2.5-7b-instruct"} \\ 
&  "learning\_rate": 5.0e-6, "num\_train\_epochs": 5.0, "bf16": true & \\
\textcolor{black}{DPO-DG} & \textcolor{black}{"temperature": 0, "max\_tokens": 1024,"train\_batch\_size": 4,"finetuning\_type": lora, } & \textcolor{black}{model = "qwen-2.5-7b-instruct"} \\ 
&  "learning\_rate": 5.0e-6, "num\_train\_epochs": 5.0, "bf16": true &\\
\textcolor{black}{DPO-MAG} & \textcolor{black}{"temperature": 0, "max\_tokens": 1024,"train\_batch\_size": 4,"finetuning\_type": lora, } & \textcolor{black}{model = "qwen-2.5-7b-instruct"} \\ 
&  "learning\_rate": 5.0e-6, "num\_train\_epochs": 5.0, "bf16": true & \\
\hline
\end{tabular}
}}
\end{table*}

\section{Supplementary Information}
This paper utilized AI tools including Google Translate for assisted translation when presenting prompts and examples, and employed the use of a Cursor for coding to enhance efficiency. No potential risks were involved in the course of this study.